\documentclass[runningheads]{llncs}
\usepackage{graphicx}
\usepackage{amsmath,amssymb} % define this before the line numbering.

\usepackage{color}
\usepackage{graphicx}

\usepackage[titlenumbered,ruled]{algorithm2e}
\usepackage{amsfonts}
\usepackage{algpseudocode}
\usepackage[dvipsnames]{xcolor}
\usepackage{caption}
\usepackage{subcaption}
\usepackage{verbatim}
\usepackage{array}
\usepackage{soul}
\usepackage{booktabs}
\usepackage{adjustbox}
\usepackage{tabularx,ragged2e,booktabs}
\usepackage{tabu}
\usepackage{float}
\usepackage{hyperref}
\begin{document}
\SetKwComment{Comment}{$\triangleright$\ }{}

\pagestyle{headings}
\mainmatter
\def\ECCV16SubNumber{***}  % Insert your submission number here

\color{blue}
\title{Variability Matters : Evaluating inter-rater variability in histopathology for robust cell detection} % Replace with your title
\color{black}

\titlerunning{Variability Matters}

\author{Cholmin Kang\inst{1}\orcidID{0000-0001-6321-0003} 
Chunggi Lee\inst{1}\orcidID{0000-0002-6164-2563} \and 
Heon Song\inst{1}\orcidID{0000-0003-3435-0271} \and 
Minuk Ma\inst{1}\orcidID{0000-0003-0416-8479}\and 
S\'ergio Pereira\inst{1}\orcidID{0000-0002-4298-0903}}

\authorrunning{Kang et al.}

\institute{Lunit, Seoul 06241, Republic of Korea 
\email{kcholmin@gmail.com,cglee@lunit.io,heon.song@lunit.io,\\
akalsdnr3@gmail.com,sergio@lunit.io}\\
\url{http://lunit.io} }

\maketitle

% https://www.overleaf.com/learn/latex/Using_colours_in_LaTeX
\newcommand{\authorcomment}[3]{\noindent\textsf{\textcolor{#1}{[\textbf{#2:} \textit{#3}]}}}
\newcommand{\dgyoo}[1]{\authorcomment{cyan}{Donggeun}{#1}}
\newcommand{\cglee}[1]{\authorcomment{orange}{cglee}{#1}}
\newcommand{\hsong}[1]{\authorcomment{blue}{Heon}{#1}}
\newcommand{\minuk}[1]{\authorcomment{green}{Minuk}{#1}}
\newcommand{\smp}[1]{\authorcomment{teal}{S\'ergio}{#1}}
\newcommand{\steve}[1]{\authorcomment{red}{Cholmin}{#1}}

\begin{abstract}

Large annotated datasets have been a key component in the success of deep learning. 
However, annotating medical images is challenging as it requires expertise and a large budget. 
In particular, annotating different types of cells in histopathology suffer from high inter- and intra-rater variability due to the ambiguity of the task.
Under this setting, the relation between annotators' variability and model performance has received little attention. 
We present a large-scale study on the variability of cell annotations among 120 board-certified pathologists and how it affects the performance of a deep learning model.
We propose a method to measure such variability, and by excluding those annotators with low variability, we verify the trade-off between the amount of data and its quality.
We found that naively increasing the data size at the expense of inter-rater variability does not necessarily lead to better-performing models in cell detection. 
Instead, decreasing the inter-rater variability with the expense of decreasing dataset size increased the model performance. 
Furthermore, models trained from data annotated with lower inter-labeler variability outperform those from higher inter-labeler variability. 
These findings suggest that the evaluation of the annotators may help tackle the fundamental budget issues in the histopathology domain

\end{abstract}

\section{Introduction}

Histopathology plays an important role in the diagnosis of cancer and its treatment planning. The process, typically, requires pathologists to localize cells and segment tissues in  whole slide images (WSIs) \cite{diao2021human}. However, with the advent of digital pathology, these tedious and error-prone tasks can be done by Deep Learning models. Still, these models are known to require the collection of large annotated datasets. Particularly, cell detection needs annotations from very large numbers of individually annotated cells.

\begin{figure}[!t]
\centering
    \includegraphics[width=0.95\textwidth]{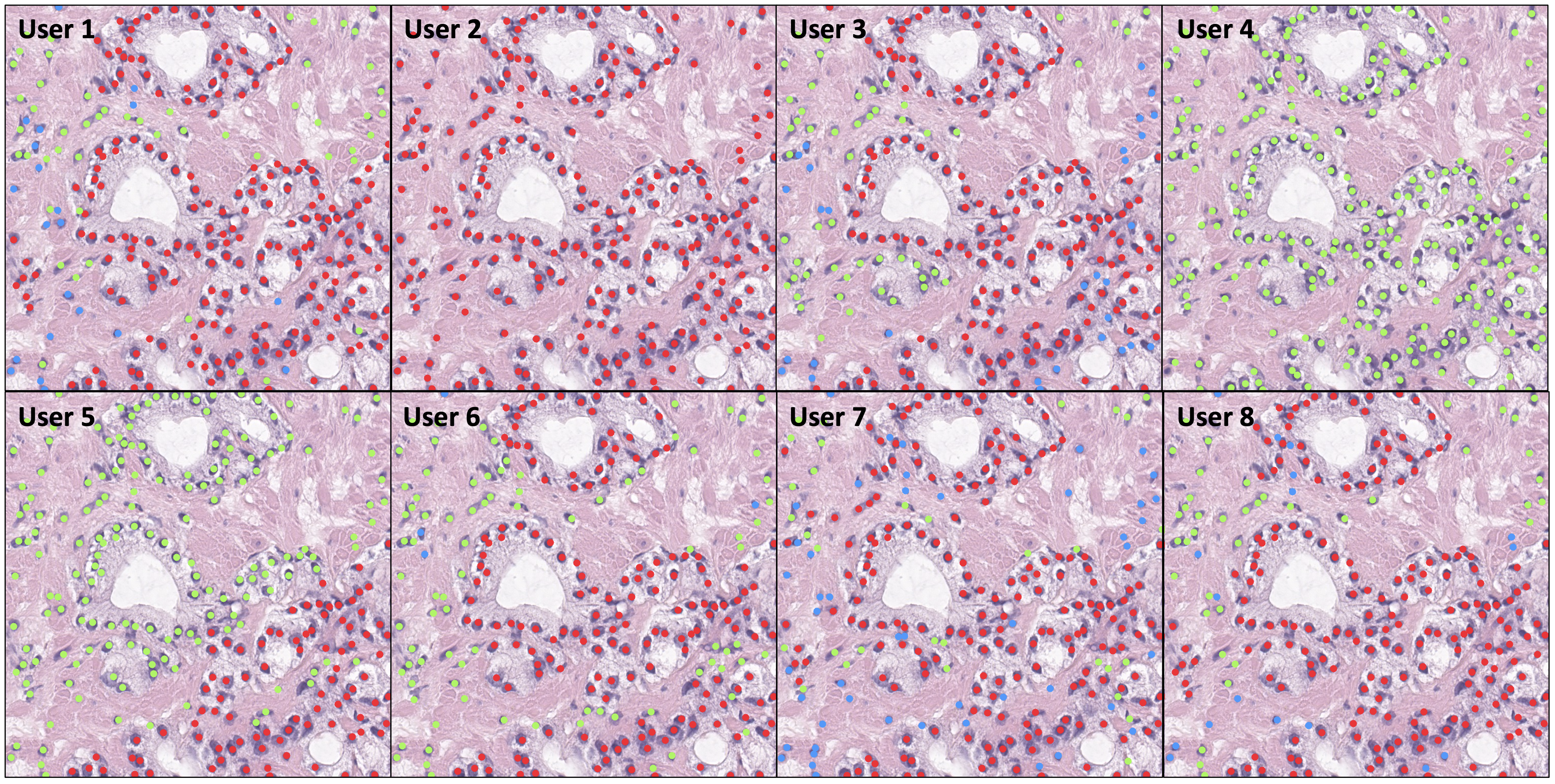}

    \caption{Example of inter-rater variability in the annotation of cells. The colored circles represent: {\color{OrangeRed}red} - tumor cell, {\color{RoyalBlue}blue} - lymphocytes, and {\color{LimeGreen}green} - other cells.}
    \label{fig:varianceexample}
\end{figure}

The identification of cells in WSIs is challenging due to their diversity, subtle morphological differences, and the astounding amount of cells that exist in a single WSI.
Therefore, when manually performed, this task is time-consuming, and suffers from high inter- and intra-rater variability among annotators \cite{annotatingHard1,annotatingHard2}. 
For example, \autoref{fig:varianceexample} shows the large discrepancy in cell annotations among different annotators.

The inconsistencies among the annotators negatively affect the performance of the model \cite{crowdisnoisy}. 
For example, when using such inconsistent data, the model may overfit to the noise and generalize poorly \cite{trainingFromNoisyData}. 
Therefore, some degree of consistency in the annotation must be guaranteed. 

To improve the consistency among annotators and reduce the variability due to annotations in the dataset, we propose to exclude annotators with large variability from other annotators. This method is based on the existence of an anchor annotator who works as a reference point to which each annotator are compared. The concept of anchor has been recently proposed \cite{wang2020systematic,miller2021class}, where a cluster center is fixed for more efficient and stable training. We propose to have an anchor annotator that annotated a control set. In this way, we can measure the conformity of the other annotators in relation to the anchor one. During annotation, a given expert can receive a new image to annotate, or, instead, with some probability a control image. The set of annotated control images of an annotator is used to measure its conformity to the anchor annotator through a modified F1-score (mF1). In this way, annotators with high conformity would achieve lower variability, thus, more reliable annotations in relation to the anchor.

We collected and annotated a total of 29,387 patches from 12,326 WSIs. 
A total of 120 board-certified pathologists participated in the annotation process, which, to the best of our knowledge, is the largest-scale pathology annotation analysis reported in the literature.  
Based on our conformity, we divide the annotated data and perform an analysis on the discrepancy among annotators.
Our results show the conformity of 0.70, with a standard deviation of 0.08.
Furthermore, the kappa score showed a mean agreement of $\kappa$=0.43 with a minimum value between $\kappa$=-0.01 and $\kappa$=0.87. 
These results show relatively low conformity among most of the annotators. Furthermore, 
we also verify that the data with lower variability leads to better-performing models. 
Indeed, the model trained with a smaller dataset with lower variability showed better performance when compared to the model trained with the full dataset. 
This supports the hypothesis that a smaller higher quality dataset can lead to a well-performing model, at a reduced cost, when compared with a large dataset with high inter-rater variability.

The contributions of this paper are five-fold. 
1) We propose a method for calculating the conformity between annotators and an anchor annotator. In this way, we can measure the variability among annotators in cell detection tasks in histopathology.
2) We observe that the annotations created by pathologists have high inter-rater variability based on our conformity method.
3) By employing the before-mentioned conformity method, we propose a process to reduce annotation variability within a dataset by excluding annotators with low conformity in relation to an anchor. This leads to more efficient usage of the annotation budget, without sacrificing performance. 
4) We show that a dataset with low inter-rater variability leads to better model performance.
5) To the best of our knowledge, this is the first work that tackles the problem of measuring variability among annotators in a large-scale dataset for a cell detection task, with the participation of a large crowd of board-certified pathologists.

\color{black}
\section{Related work}

Deep Learning techniques, particularly Convolutional Neural Networks (CNN) revolutionized the field of computer vision and image recognition \cite{imageNet,girshick2014rich} by achieving unprecedented performances. As such, it has also been extensively explored in the domain of digital pathology \cite{CellComputer}. Cell detection tasks have received attention due to its laborious and error-prone nature \cite{chen2014deep,gao2016hep,cirecsan2013mitosis,xue2017cell}. However, there has been little attention to the effect of inter-rater variability in cell detection tasks, and how it affects the performance of the models.

In comparison to natural images, the annotation of medical images is a more challenging task. First, domain experts are needed, e.g., medical doctors. Second, due to the difficulty and ambiguity of the tasks, these tasks are more prone to subjectivity which leads to the disagreement between experts \cite{hekler2019deep,bertram2022computer}. The inter-rater variability and reproducibility of annotations have been studied in some medical-related tasks, such as sleep pattern segmentation in EEG data \cite{schaekermann2019understanding}, or segmentation in Computed Tomography images \cite{haarburger2020radiomics}. In the case of histopathology, some studies demonstrate that computer-assisted reading of images decreases the variability among observers \cite{hekler2019deep,bertram2022computer}.

To overcome the disagreements, some works attempt to utilize the annotations of multiple raters by learning several models from single annotations and from the agreement of multiple annotators \cite{sudre2019let}. In this way, the effect of disagreement can be mitigated, but it assumes multiple annotations for the same images. Other approaches accept that inter-rater variability will exist, and explore crowds of non-experts to collect large amounts of data by crowd-sourcing, either with or without the assistance of experts. Nonetheless, the scopes were limited to tasks that do not require a high degree of expertise or the annotation of many instances  \cite{canmassesofnonexperts,DoWeNeed}. Such assumptions do not hold in cell annotation in histopathology. Crowd-sourcing cell annotations were previously performed \cite{amgad2021nucls}, but, the evaluation of the conformity of the annotators in this setting, where there is high inter-rater variability, remains unexplored. Inconsistencies among the annotators introduce noise to the annotations \cite{crowdisnoisy}. In turn, it can lead to models that overfit to the noise and generalize poorly \cite{trainingFromNoisyData}. Therefore, some degree of label variability must be guaranteed.

Other related prior work tackles the problem of lack of conformity between annotators from the perspective of a noisy label  \cite{smallOracleDataset,mentornet}, or, tried to classify the reliability of annotators using the probabilistic model\cite{yan2014learning}. Other  works tried to resolve large variability of annotator performance using the EM algorithm\cite{raykar2010learning} or Gaussian process classifiers \cite{rodrigues2014gaussian}. However, these approaches rely on the training dynamics, so, it is not possible to distinguish if a given sample is noisy or just difficult. Moreover, the noisy data is discarded, raising two concerns: 1) resources are used to gather unused data, and 2) the dataset needs to be sufficiently large to allow discarding data without impacting the performance of the model. Instead, we hypothesize that noise can be introduced by annotators with high variability to other annotators, and preventive selection of conforming annotators leads to higher quality annotations. This becomes even more necessary in large-scale settings, where a large pool of annotators needs to be sourced externally.

\color{black}
\section{Evaluating annotators and variability}

During the annotation process, the annotators receive unlabeled patches to annotate from the training set $\mathbf{T}$. Control set $\mathbf{C}$ refers to the unlabeled dataset that will later be used to measure the variability among annotators. Images from the control set $\mathbf{C}$ are randomly, and blindly, assigned to the annotators.
% with a probability of \textit{p}\%. In this experiment, we used 0.03\%-0.08\%. 
Since the annotators do not know when a control image is provided, this ensures that the annotation skill showed in $\mathbf{C}$ is similar to that of $\mathbf{T}$. Similar methods are used in practice when evaluating crowd-sourced workers~\cite{randomassigntest}. An \textit{anchor annotator} is chosen as the reference annotator to perform analysis on discrepancy (i.e., variability) among annotators. The annotations of each annotator are compared with the annotations done by the \textit{anchor annotator}. Conformity is measured between each annotated cell of an annotator and the annotated cells of an \textit{anchor annotator}. The conformity is used to show the variability of annotators and annotators with large conformity will have small variability among them.

\begin{figure}[!h]
\centering
    \includegraphics[width=0.95\textwidth]{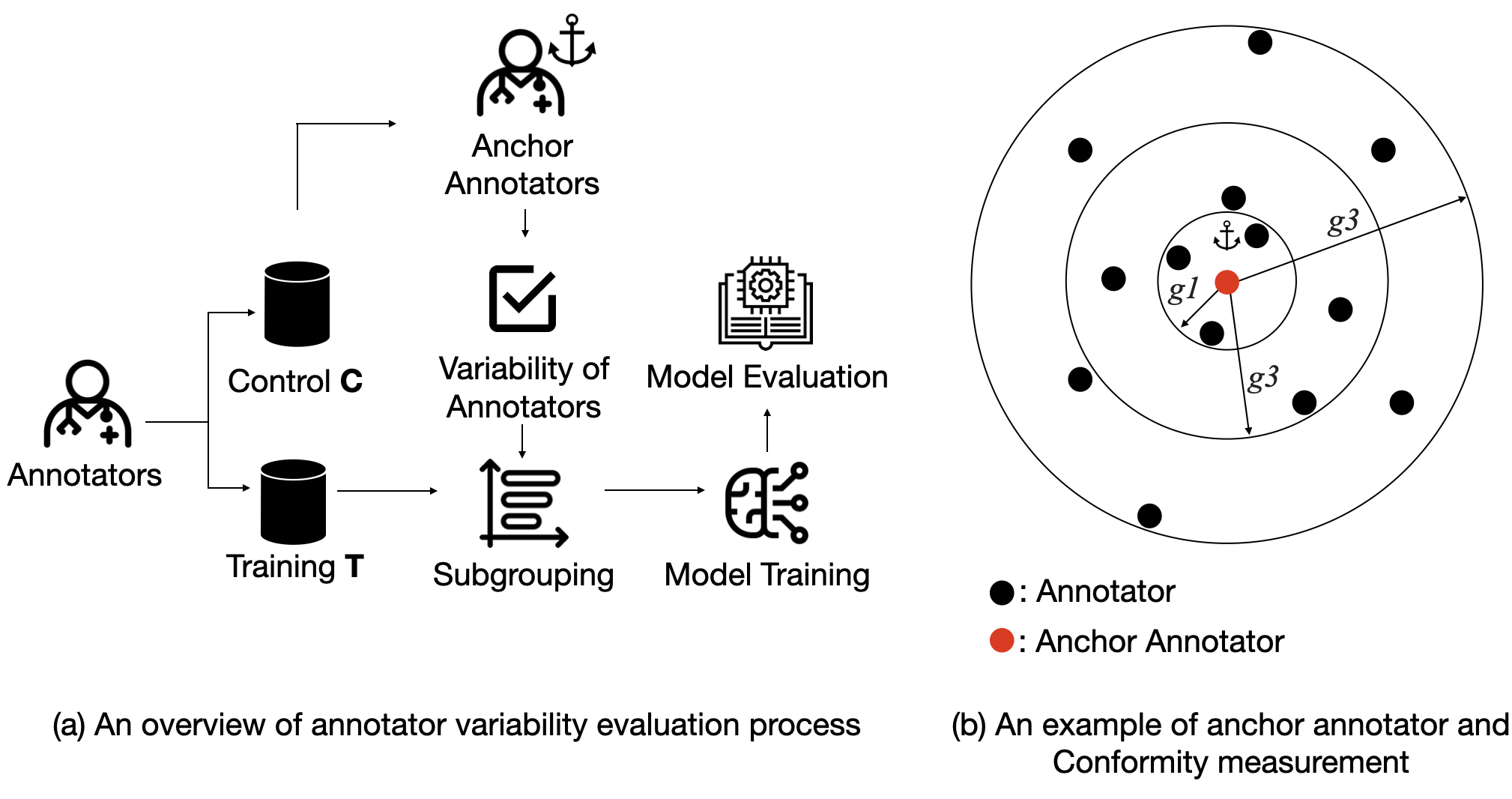}
    
    \color{blue}
    \caption{Evaluation of inter-rater variability. (a) The control set is created to blindly evaluate the annotator conformity during the annotation process. A score is computed per annotator, which is used to evaluate the impact of their annotations in the cell detection task. (b) Annotator selection method. Conformity is measured between each annotation done by an annotator and anchor annotator. In the annotator selection phase, annotations are divided into subgroups (such as $g_1$, $g_2$, or $g_3$ in the figure) according to annotator conformity percentiles.
    }
    \color{black}
    \label{fig:overview}
\end{figure}

Finally, annotators are divided into groups according to their conformity score. The model is, then, trained on the data of each group and aggregation of groups. The annotation, conformity measurement, and model development process are depicted in \autoref{fig:overview}(a).

\subsection{Anchor annotator}\label{anchor_annotator}
\begin{figure}[!h]
\centering
    \includegraphics[width=0.95\textwidth]{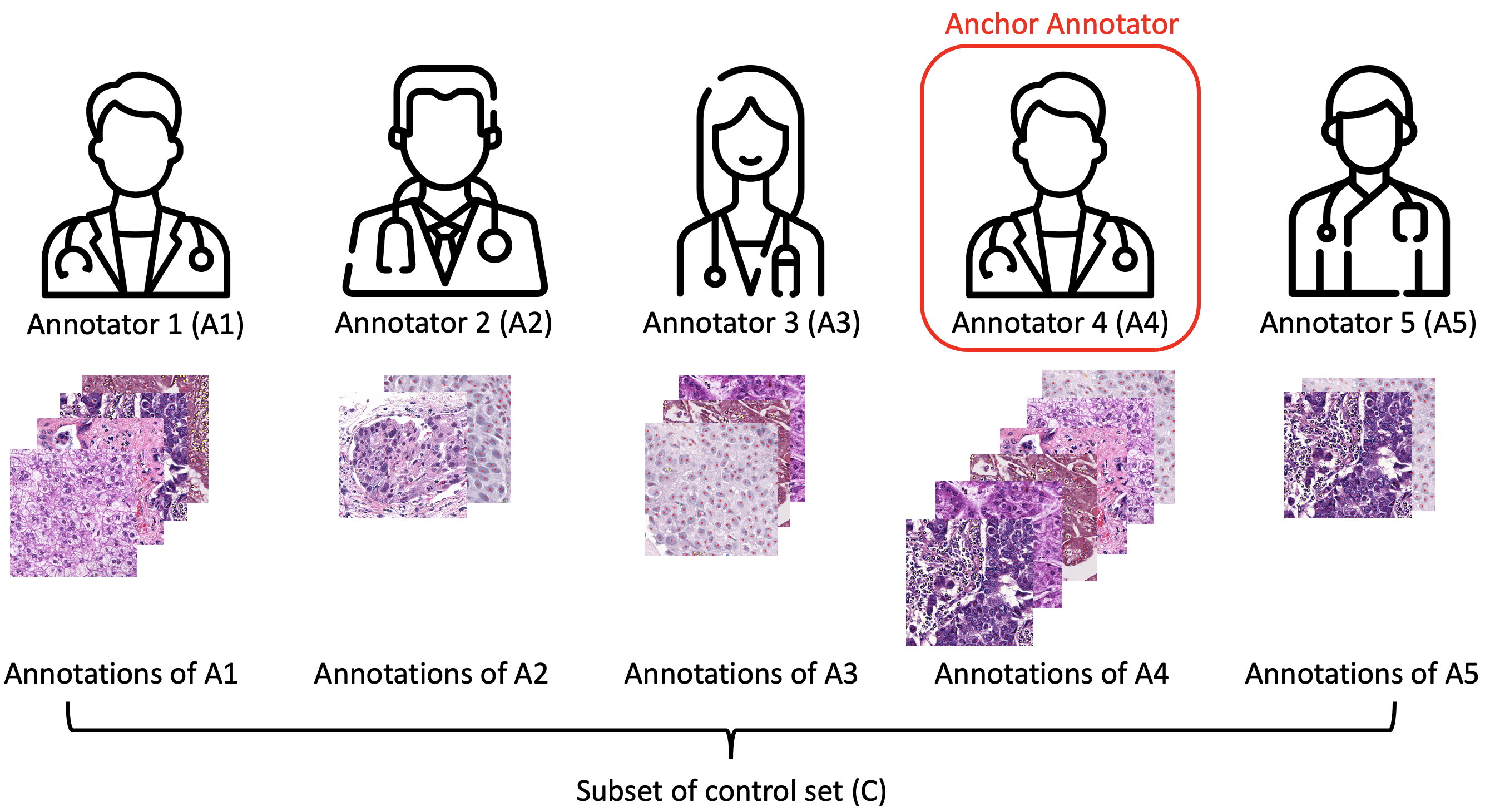}
    \caption{
    A conceptual overview of anchor annotator selection. Anchor annotator is chosen among annotators with the most amount of annotations done in control set $C$. As shown in the figure, the patches annotated by A4 is the superset of patches annotated by A1, A2, A3, and A5. Thus, A4 is chosen as an anchor annotator.
    }
    \label{fig:anchorannotator}
\end{figure} 
In order to perform an analysis of the variability among annotators, reference data should be set to measure how much each annotator is far from the reference point. In this work, one annotator is defined as the anchor annotator and is chosen as a criterion to evaluate variability among annotators. The rationale for setting an anchor annotator and evaluating variance is motivated by the concept of an anchor cluster center \cite{wang2020systematic,miller2021class}, where a cluster center is fixed for more efficient and stable training. The conceptual overview of anchor annotator selection is shown in \autoref{fig:anchorannotator}. As shown in the Figure, an anchor annotator is the one with most number of images annotated so that the annotated images are the superset of the patches annotated by other annotators.

\subsection{Annotator variability and conformity calculation}

\begin{algorithm}[!h]
\small
\caption{Conformity measurement algorithm for an individual annotator}
\KwIn{
$\mathbf{A}$ and $\mathbf{C}$: sets of annotations $\mathbf{A}_n^k=\left\{\mathbf{a}_{n,1}^k,\ldots\right\}$ and $\mathbf{C}_n^k=\left\{\mathbf{c}_{n,1}^k,\ldots\right\}$  where $\mathbf{a}_{n,p}^k$ and $\mathbf{c}_{n,q}^k$ indicates the $p$-th and $q$-th annotations of the $n$-th control patch and class $k$ \newline

$\theta$: distance threshold
}
\KwOut{$conf$: conformity}
$\mathbf{s} \gets \mathbf{0}$ \Comment*[r]{Initialize a vector indicating F1-score for each class}
\ForEach{class $k$}{
$TP \gets 0$, $FP \gets 0$, and $FN \gets 0$ \Comment*[r]{Initialize TP, FP, and FN}
\ForEach{$n$-th control data patch}{
$G_A \gets \emptyset$ and $G_M \gets \emptyset$ \Comment*[r]{Initialize assigned index sets}
$D \in \mathbb{R}^{|\mathbf{A}_n^k| \times |\mathbf{C}_n^k|} \gets \mathbf{0}$ \Comment*[r]{Initialize the pairwise distance matrix}
\ForEach{index $p$ and $q$}{
$D_{p,q} \gets \| \mathbf{a}_{n,p}^k - \mathbf{c}_{n,q}^k \|_2$ \Comment*[r]{Compute the euclidean distance}
}
\ForEach{$i$-th index pair $(p^*_i, q^*_i)$ such that $\forall j > i, D_{p^*_i,q^*_i} \le D_{p^*_{j},q^*_{j}}$}{
\If{$D_{p^*_i,q^*_i} \le \theta$, $p^*_i\notin G_A$, and $q^*_i\notin G_M$}{
$G_A \gets G_A \cup \{p^*_i\}$ and $G_M \gets G_M \cup \{q^*_i\}$ \Comment*[r]{Assign indices}
}
}
$TP \gets TP + |G_A|$ \Comment*[r]{Update TP}
$FP \gets FP + |\mathbf{A}_n^k|-|G_A|$ \Comment*[r]{Update FP}
$FN \gets FN + |\mathbf{C}_n^k|-|G_M|$ \Comment*[r]{Update FN}
}
$s_k \gets \frac{2TP}{2TP + FP + FN}$ \Comment*[r]{Compute the F1-score for class $k$}
}
$conf \gets \frac{1}{|\mathbf{s}|}\sum_k s_k$ \Comment*[r]{Compute the conformity as mean F1-score}
\label{alg:scoring}
\end{algorithm}
Annotator conformity is calculated using \autoref{alg:scoring}.
To compute the annotators' conformity, a hit criterion needs to be defined. We consider a circular region of radius $\theta=10$ micrometers centered in each cell of $\mathbf{C}$ in a given patch. Annotations from $\mathbf{C}$ that are located inside a given circle are True Positive (TP) candidates. Among these candidates, the closest point that matches the same class as the annotation from $\mathbf{C}$ is considered a TP. Given this criterion, we also compute False Positive (FP) and False Negative (FN) quantities. Based on the TP, FP, and FN, F1-score per class is computed and the conformity of the annotator is defined as mean F1-score over classes.

\subsection{Annotator grouping based on the conformity}
Finally, annotators are divided into subgroups according to their conformity to the anchor annotator.  \autoref{fig:overview}(b) shows the annotators' annotation distribution. First, an anchor annotator is selected. Then, conformity between the anchor annotator and other annotators is calculated. Finally, each annotator is evaluated by computing the conformity, as described in \autoref{alg:scoring}. In this work, we consider this conformity as a variability. If all annotations done by all annotators are equal to those done by the anchor annotator, the conformity will be 1. However, if the annotations are different from the anchor annotator, the annotator conformity calculated by \autoref{alg:scoring} will decrease, leading to increased variability and decreasing conformity. 
\color{black}

% \section{Results and Discussion}
\section{Experimental Results}

\subsection{Dataset}
\begin{table}[!t]
\caption{Description of the Control, Training, and Validation sets in terms of the number patches, annotated cells, and WSIs.} 

\centering
\begin{tabular}{c|c|c|c|c}
\toprule
Dataset           & Patches & Annotated cells & Avg annotated cells per patch & Number of WSI\\ \hline
Control set & 150           & 32,746    & 218.31 & 141                \\ 
Training set      & 21,795         & 3,727,343   & 171.02 & 8,875                \\ 
Validation set    & 7,442          & 1,390,551                     & 186.85 & 3,310                \\ \bottomrule
\end{tabular}
\label{tab:dataset}
\end{table}

\begin{figure}[!t]
\centering
    \includegraphics[width=0.95\textwidth]{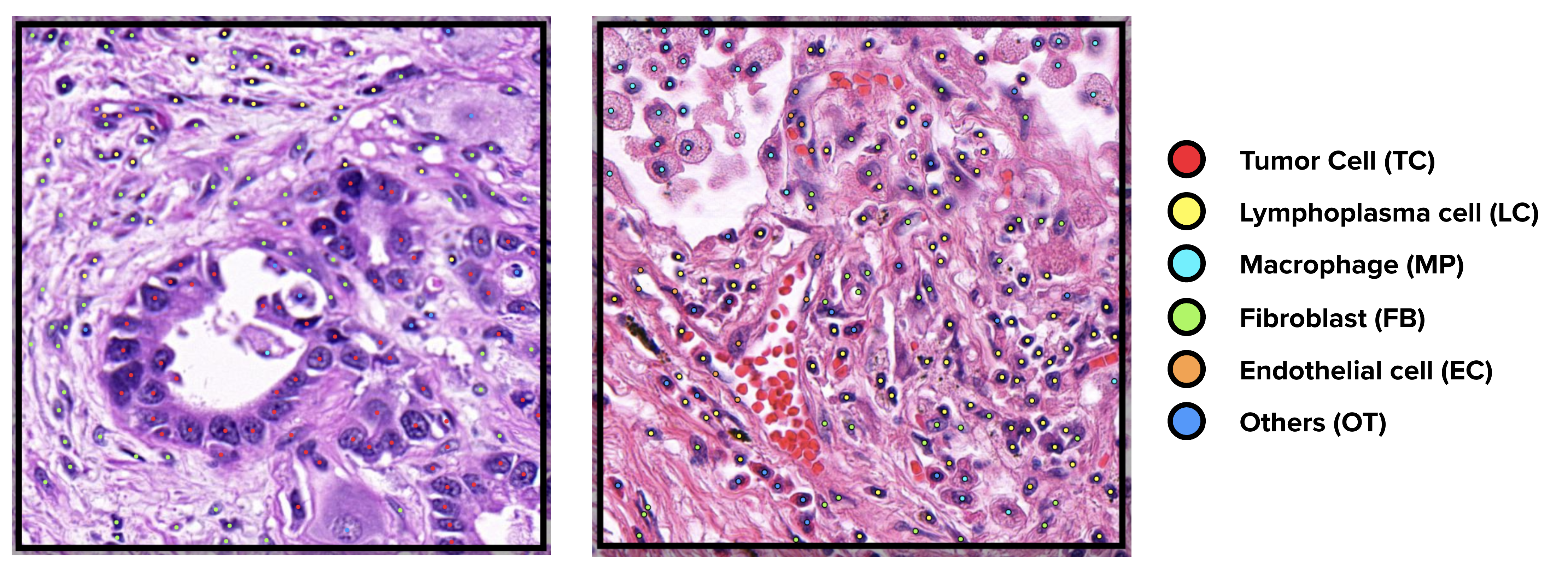}
    \caption{
    Example of the reference images provided to annotators alongside the annotation guidelines.
    }
    \label{fig:guideline}
\end{figure}

We collected a large-scale dataset for the task of tumor and lymphocyte cells detection in H\&E-stained WSIs, from 16 primary origins cancers (e.g. lung, breast, kidney, etc). As WSIs are typically large for exhaustive annotation, occupying several giga-pixels squared in area, it is a common practice to divide them into smaller patches \cite{patch}. We sampled patches of size 1024 $\times$ 1024 pixels from a total of 12,326 WSIs. The annotations consist of point annotations that locate the nuclei of cells, as well as their classes. We collect 3 sets at the WSI level: control set ($\mathbf{C}$),
% \color{blue}
Training set ($\mathbf{T}$)
\color{black}
and Validation set ($\mathbf{V}$). The validation set is annotated by three annotators, and has no overlap with $\mathbf{C}$.
The number of images and cell annotations in each set is presented in \autoref{tab:dataset}. A total of 29,387 unique patches are annotated. As can be seen in the Table, the average number of annotated cells in a single image is similar among $C$, $T$, and $V$. A large crowd of 120 board-certified pathologists participated in the annotation process. During the annotation process, annotators are provided with guidelines on how to annotate each cell. Annotators are guided to annotate each cell to one of six classes: Tumor Cell, Lymphoplasma cells, Macrophage, Fibroblast, Endothelial cell, and others. The class "other" include nucleated cells with vague morphology that are not included in the specific cell types. Each cell is explained with its definition and its clinical characteristics. Furthermore, visual examples were provided to exemplify the guides, such as in \autoref{fig:guideline}.
\subsection{Cell detection method and experimental details} 

Having the conformity scores of the annotators, we evaluate its efficacy in filtering out annotations from annotators with higher variability by training a model in the downstream task of cell detection. Similar to \cite{swiderska2019learning}, we pose the problem as a dense pixel classification task, but we use a DeepLabv3+ \cite{deeplab} architecture. The point annotation of each cell is converted to a circle of radius of 5 pixels, hence, resulting in a map similar to semantic segmentation tasks. The output is a dense prediction likelihood omap that, similar to \cite{swiderska2019learning}, requires a post-processing stage to retrieve the unique locations of the cells\footnote{Post-processing consists of Gaussian filtering ($\sigma=3$) the likelihood maps, followed by local maxima detection within a radius of 3 pixels.}. 
In this experiment, we target Tumor-Infiltrating Lymphocyte (TIL) task which requires the number of tumor cells and lymphocytes in given WSI. Therefore, among the annotations, we regard macrophage, fibroblast, and endothelial cell as "others" class. 
All models are trained using the Adam \cite{adam} optimizer (learning rate of 1e-4) and the soft dice loss function \cite{milletari2016v}. Experiments are performed on 4 NVIDIA V100 GPUs, and implemented using PyTorch (version 1.7.1) \cite{pytorch}.

\subsection{Conformity of the annotators}

\begin{figure}[!ht]
\centering
    \includegraphics[width=0.99\textwidth]{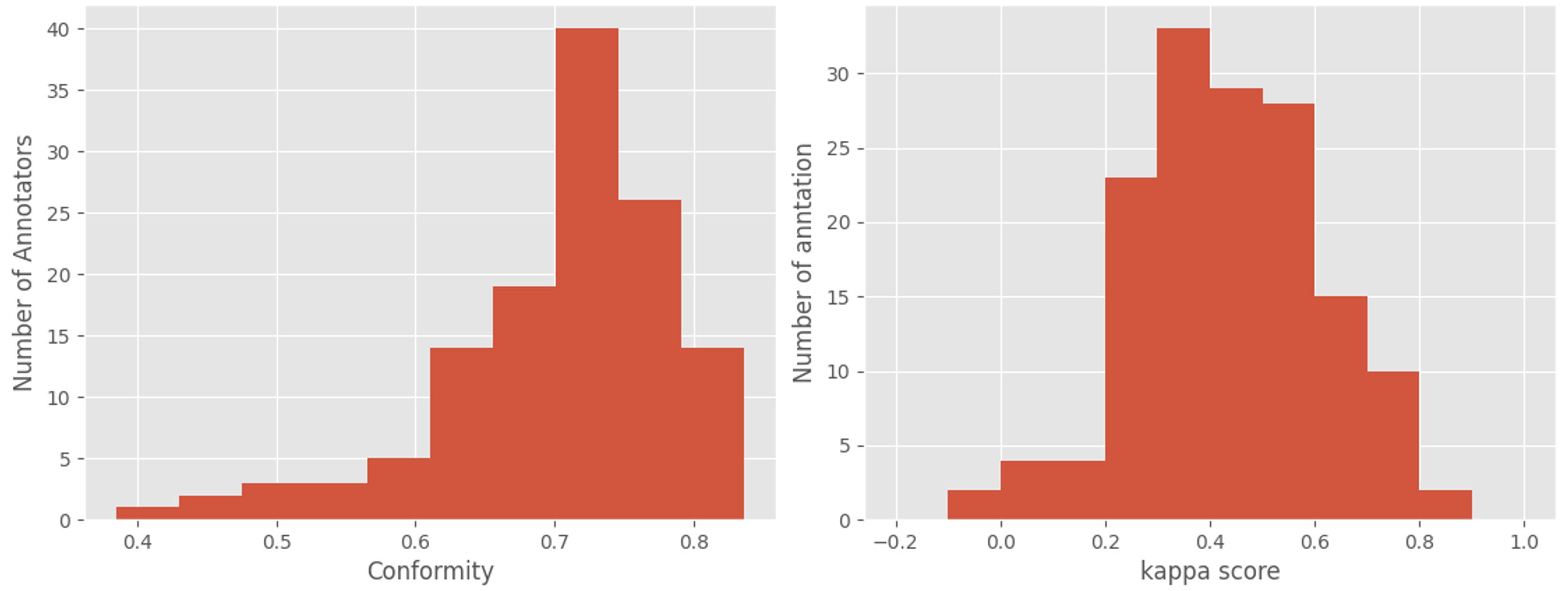}

    \caption{Distributions of the variability of the external annotators as mF1-score (left), and the agreement of the annotations by the external annotators in relation to the set $\mathbf{C}$, measured in terms of Fleiss' kappa (right). } 
    % \smp{Maybe the right figure has too many bins, which makes too small values on the y-axis.}}
    \label{fig:scoreDistr} 
\end{figure}

Following the proposed evaluation procedure of the annotators, we computed the individual annotator conformity. In this work, an annotator with the most number of annotations done from set $C$ is chosen as an anchor annotator so that most annotators can have conformity calculated. The closer a score is to 1, the better the annotator conforms with an anchor annotator. \autoref{fig:scoreDistr}, left, shows the distribution of the conformity score. The median conformity is 0.70 (minimum=0.38, and maximum=0.83). We further conduct a Shapiro-Wilk Normality test \cite{shapiro} and verify that it does not follow a Normal distribution at a significance level $\alpha=0.05$, with a p-value$=9.427e-8$. Moreover, we observe a negative skewness of $-1.3158$, which confirms the observation of the longer left tail.

Varying expertise is a known problem when annotating cells in histopathology. Nonetheless, one would expect the conformity of the experts to reference annotation to be normally distributed. Instead, we observe that while the median annotator conformity is relatively high, there are some annotators on the left tail that perform significantly worse than the majority. This supports the observation of high inter-rater variability.

To further analyze the inter-rater variance, we measured the agreement among all annotators on each Control patch using Fleiss' kappa \cite{falotico2015fleiss}, which is a statistical measure of the reliability of agreement for categorical ratings with more than two raters. 
% \color{blue}
This test was run to show inter-rater variability rather than to show absolute quality of annotators.
\color{black}

In this analysis, the inter-rater agreement is measured in relation to three categories: tumor cells, lymphocytes, and unmatched annotations. As previously described, the annotated cells from each pathologist have been matched to the nearest cell of the same class in $\mathbf{C}$, following the hit criterion.
On the other hand, if the hit criterion is not satisfied, we treat it as an unmatched annotation.
We compute the Fleiss' kappa for all patches in $\mathbf{C}$.
The mean agreement among the annotations of the pathologists was $\kappa$=0.43 with the minimum and maximum of $\kappa$=-0.01 and $\kappa$=0.87, respectively. This range (-0.01 to 0.87) of kappa values suggest poor to an excellent agreement, but the mean is located in a fair to moderate agreement range. This further reinforces the observation that there exists high inter-rater variability. 
We observe that the classification of the cells is sensitive in specific patches and pathologists, as shown in \autoref{fig:varianceexample}.

\subsection{Impact of the conformity of the annotators in cell detection}

\begin{table}[t]
\small

\caption{Model performance by each cancer primary origin and annotation conformity percentile range. For example, the column ``$p_{0-25}$'' denotes the training set with patches annotated by annotators with conformity in percentiles 0 to 25, i.e., the 25\% annotators with the lowest conformity. At the bottom, it is presented the total annotation time in hours.}
  \centering

\begin{tabular}{>{\centering}m{3cm}|>{\centering}m{1.2cm}>{\centering}m{1.2cm}|>{\centering}m{1.2cm}>{\centering}m{1.2cm}|>{\centering}m{1.2cm}>{\centering\arraybackslash}m{1.2cm}}

\toprule

& \multicolumn{6}{c}{Percentile ranges} \\ \hline
Primary Origin                           &                     

$p_{0-25}$ & $p_{75-100}$ & $p_{0-50}$ & $p_{50-100}$ & $p_{25-100}$ & $p_{0-100}$ \\ \hline
\multicolumn{1}{c|}{Biliary Tract}       & {62.09}  & \multicolumn{1}{c|}{62.9} & {63.71} & {64.06}  & {65.1} & {64.89} \\
\multicolumn{1}{c|}{Breast}              & {62.32} & \multicolumn{1}{c|}{64.74} & {64.53} & {65.53} & {64.83} & {65.3} \\
\multicolumn{1}{c|}{Colorectum}          & {61.87} & {63.09} & {63.83} & {63.8} & {64.02 }&{63.84} \\
\multicolumn{1}{c|}{Esophagus}           & {64.11} & {61.42} & {65.56} & {66.24} & {66.33} & {66.19} \\
\multicolumn{1}{c|}{Head and Neck}       & {59.07} & {61.68} & {61.76} & {62.54} & {62.63} &{62.61} \\
\multicolumn{1}{c|}{Kidney}              & {49.68} & {53.94} & {52.10} & {56.07} & {57.34} &{56.49} \\
\multicolumn{1}{c|}{Liver}               & {62.50} & {65.16} & {64.06} & {66.09} & {65.29} &{65.16} \\
\multicolumn{1}{c|}{Lung}                & {58.18} & {60.24} & {60.27} & {61.24} & {60.75} &{60.72} \\
\multicolumn{1}{c|}{Melanoma}            & {59.08} & {60.03} & {60.18} & {61.64} & {61.73} &{61.27} \\
\multicolumn{1}{c|}{Ovary}               & {53.92} & {55.64} & {56.57} & {57.56} & {57.79} &{57.32} \\
\multicolumn{1}{c|}{Pan Urinary}         & {62.71} & {64.82} & {65.33} & {65.45} & {65.02} &{64.8} \\
\multicolumn{1}{c|}{Pancreas}            & {56.68} & {58.49} & {58.62} & {59.93}& {61.16 }&{60.06} \\
\multicolumn{1}{c|}{Prostate}            & {55.70} & {56.41} & {56.12} & {58.13} & {58.39} &{58.64} \\
\multicolumn{1}{c|}{Stomach}             & {59.74} & {61.07} & {62.13} & {62.79} & {63.54} &{63.10} \\
\multicolumn{1}{c|}{Uterine Cervix}      & {61.86} & {64.33} & {64.37} & {65.23} & {65.69} &{65.64}\\
\multicolumn{1}{c|}{Uterine Endometrium} & {53.18} & {54.70} & {54.13} & {56.07} & {55.53} &{55.51}\\
\hline
\multicolumn{1}{c|}{Average mF1}    & {58.92} & 60.77 & 60.87 & 62.03 & 62.18 &61.76\\ \hline

\multicolumn{1}{c|}{Annotation time (hours)}    & {598} & 575 & 1,122 & 1,176 & 1,703 & 2,298\\

\bottomrule
\end{tabular}%

\label{tab:sameSize}%
\end{table}%

In this experiment, we divide the annotators by their conformity, from lower to higher. Following this conformity, we define 4 subsets of the data based on the percentile of the conformity of the annotators. Therefore, the interval $p_{0-25}$ contains 25\% of the patches in the dataset that were annotated by annotators with high variability. The intervals $p_{25-50}$, $p_{50-75}$, and $p_{75-100}$ also contain 25\% of the patches each, with decreasing annotator variability as compared to the anchor annotator. Therefore, following our hypothesis, the interval $p_{75-100}$ must contain the annotations with the lowest variability. We hypothesize that data annotated by annotators with lower variability should lead to better-performing cell detection models. \autoref{tab:sameSize} shows the performance of the cell detection model in terms of conformity by the before-mentioned intervals of data across the 16 cancer primary origin organs. The results are the average of 3 models trained with different random seeds. We use the paired Wilcoxon Signed Rank test~\cite{woolson2007wilcoxon} at a significance level $\alpha=0.05$ to compare results and test our hypothesis.

We compare the performance of models trained with similar amounts of data, but with different conformity in relation to the annotation. First, we contrast interval $p_{0-25}$ and $p_{75-100}$, i.e., the data annotated by the highest variance and lowest variance annotators, respectively. Note that both sets contain 25\% of patches of the full training dataset. We observe that the data from the best performing annotators (interval $p_{75-100}$) leads to better performance compared to interval $p_{0-25}$. This is true for all of cancer primary origins and is corroborated by an existing statistical difference between the results (p-value=0.00614). Similarly, when we use 50\% of the data to train the cell detection model (intervals $p_{0-50}$ and $p_{50-100}$), a similar trend is observed. The model trained from the data from annotators with lower variability outperforms the model trained with data from interval $p_{0-50}$; again, a statistical difference was found (p-value=0.00054). This result shows that given the same amount of data, the model trained with high conformity data outperforms that trained with low conformity data.

% \color{blue} 
Additionally, we evaluate the performance of models trained from smaller datasets labeled by annotators with lower variability compared to the use of the full training set. The intervals $p_{25-100}$ and $p_{50-100}$ contain an upper 75\% and 50\% of the training dataset, respectively. Nonetheless, in both cases, they outperform the model trained with the full training set, with statistical significance in the case of the interval $p_{25-100}$ (p-value=0.01778). These results suggest that smaller but low inter-rater variability datasets can outperform models trained from the larger datasets, but with higher variability annotations. 

The data from the interval $p_{50-100}$ represent the upper 50\% of the available training data, but the performance is on par with the variability setting of the interval $p_{25-100}$. Therefore, in retrospect, we could have collected 50\% fewer data, which would have reduced the total amount of annotation time from 2298 hours to 1176 hours. This is especially significant as the annotation process requires expert pathologists. 
% \color{blue}
Thus, we believe that the proposed annotation scoring procedure can be helpful in using the available budget more effectively, by requesting annotations from the experts with low variability to anchor annotator and obtain either a smaller dataset with good model performance or a large-scale dataset with lower inter-rater variability annotations in the future data collection process.
\color{black}

\section{Conclusions}

In this paper, we proposed a simple, yet effective method to assess the variability of a crowd of 120 experts in the time-consuming and difficult task of cell annotation in histopathology images stained with H\&E.  Our findings support the previous observations that this task suffers from high inter-rater variability. Furthermore, we conducted a proof-of-concept experiment to show the effect of inter-rater variability on a machine learning model for the task of cell detection. From the experiment, we showed that with the same size of data, the data labeled by annotators with lower inter-rater variability led to better model performance than the data labeled by annotators with higher inter-rater variability. Furthermore, excluding the data with the most inter-rater variability from the full dataset was beneficial, despite resulting in a smaller dataset. Hence, contrary to common findings in deep learning research, the fewer amount of data with lower inter-rater variability resulted in better-performing machine learning model. These findings are useful in the medical domain where annotations are costly and laborious. With this finding, we can argue that collecting fewer data from low inter-rater variability is more beneficial than collecting data without considering the inter-rater variability. Despite the findings that increasing conformity leads to increasing model performance, we recognize that there are some limitations in our work. First, our work is based on the premise that the anchor annotator is well-performing. The work is based on the assumption that all annotators are sufficiently expert and have done their best to annotate. Further investigation is needed with iterated experiments on setting annotators as anchor annotators. Second, the generalizability of our work on stained WSI other than H\&E stain is not guaranteed. Further investigation on other staining methods with more diverse skill backgrounds of digital pathologist annotators is needed. We hope the idea of scoring the annotators helps improve the annotation budget management in the histopathology domain. 
\color{black}

\clearpage

\bibliographystyle{splncs}
\bibliography{egbib}
\end{document}